\documentclass[conference]{ieeeconf}
\IEEEoverridecommandlockouts
\usepackage{cite}
\usepackage{amsmath,amssymb,amsfonts}
\usepackage{algorithmic}
\usepackage{graphicx}
\usepackage{textcomp}
\usepackage{xcolor}
\usepackage[utf8]{inputenc} 
\usepackage[T1]{fontenc}    
\usepackage{hyperref}       
\usepackage{url}            
\usepackage{booktabs}       
\usepackage{amsfonts}       
\usepackage{nicefrac}       
\usepackage{microtype}      
\usepackage{lipsum}
\usepackage{graphicx}
\usepackage{color}
\usepackage{tabularx}
\usepackage{amsmath}
\usepackage{microtype}
\usepackage{subfigure}
\usepackage{booktabs} 
\usepackage{mathtools, nccmath}
\usepackage[ruled]{algorithm2e}
\usepackage{multirow}

\def\BibTeX{{\rm B\kern-.05em{\sc i\kern-.025em b}\kern-.08em
    T\kern-.1667em\lower.7ex\hbox{E}\kern-.125emX}}
\begin{document}

\title{\LARGE \bf Deep Learning based Multi-Modal Sensing for Tracking and State Extraction of Small Quadcopters}





\author{Zhibo Zhang$^{1}$, Chen Zeng$^{2}$, Maulikkumar Dhameliya$^{2}$, Souma Chowdhury$^{2}$, and Rahul Rai$^{3}$
\thanks{$^{1}$Manufacturing and Design Lab (MADLab), Department of Mechanical and Aerospace Engineering, University at Buffalo, Buffalo, NY }%
\thanks{$^{2}$Adaptive Design Algorithms, Models and Systems (ADAMS) Lab, Department of Mechanical and Aerospace Engineering, University at Buffalo, Buffalo, NY}%
\thanks{$^{3}$Corresponding author. Geometric Reasoning and Artificial Intelligence Lab  (GRAIL), Clemson University, Greenville, SC, USA
{\tt\small rrai@clemson.edu}}%
}

\maketitle

\begin{abstract}
This paper proposes a multi-sensor based approach to detect, track, and localize a quadcopter unmanned aerial vehicle (UAV). Specifically, a pipeline is developed to process monocular RGB and thermal video (captured from a  fixed platform) to detect and track the UAV in our FoV. Subsequently, a 2D planar lidar is used to allow conversion of pixel data to actual distance measurements, and thereby enable localization of the UAV in global coordinates. The monocular data is processed through a deep learning-based object detection method that computes an initial
bounding box for the UAV. The thermal data is processed through a thresholding and Kalman filter approach to detect and track the bounding box. Training and test data is prepared by combining a set of original experiments conducted in a motion capture environment and publicly available UAV image data. The new pipeline compares favorably to existing methods and demonstrates promising tracking and localization capacity of sample experiments.
\end{abstract}


\section{Introduction}

The introduction of small low-altitude and low-speed unmanned aerial vehicles (UAVs) into the airspace is dependent on the maturity of technologies associated with their operational safety. Both collision avoidance mechanisms and active alert mechanisms (that can track friendly and hostile UAVs) form important cogs of the overall safety system needed for integration of small UAVs into the airspace \cite{pham2015survey} \cite{wallace2015examining}. 
The majority of UAV-to-UAV collision avoidance techniques can be categorized into two groups: cooperative maneuvers \cite{angb, 1272619, shim2003decentralized} and non-cooperative maneuvers \cite{pomdp}\cite{pomdp2}. 
The reciprocal avoidance maneuver\cite{van2011reciprocal, alejo2014optimal,tracejournal} is a technique whose requirements sit between the stated two categories. While the reciprocal avoidance mechanism does not require any active communication (while ensuring coherent maneuvers), it is strongly dependent on the online sensing and intruder state estimation capabilities of the own-ship UAV. Sensing and state estimation of another flying object remains an important challenge in the collision avoidance and tracking research communities \cite{stevendraft}. The work in this paper is motivated by our long term research goals of enabling fully autonomous (online) reciprocal detect-sense-estimate-avoid (R-DSEA). We have already made important accomplishments in the "avoidance planning" component \cite{tracejournal,behjataviation2019,behjatidetc2019}. This paper specifically sets out to accomplish our first milestone towards the detect-sense-estimate (DSE) components -- which is to develop and test a multi-modal sensing driven pipeline for DSE. In its current nascent form, this pipeline focuses on DSE that is based on independent RGB and thermal sensors (calibrated with a 2D planar LIDAR) that are placed on a stationary platform.

We draw inspiration from the field of object detection and tracking, which is a vital subfield of computer vision and more recently of robotics and automotive.  However, the application of object detection and tracking methods in UAV detection and tracking remains limited. 
Most of the existing work use only the monocular camera as input for object detection. Use of monocular camera makes it difficult to obtain the distance information, and could also be sensitive to fast changing background and deliberate misguidance with images \cite{roelofsen2015reciprocal}. Furthermore, the majority of the existing methods relies on heuristic identification of the correct motion model or features. On the other hand, the usage of deep learning in this domain has been mostly limited to the classification of the flying object (such as UAV vs. bird, or type of UAV).

Our multi-sensor based computational pipeline leverages deep learning and image processing mechanisms to perform UAV detection and tracking. Three different sensors are utilized in our pipeline: a monocular camera, a thermal camera, and a LIDAR. We use a Faster R-CNN to process the monocular camera footage since Faster R-CNN merges detection and tracking in one step, and exhibits excellent performance in terms of speed and accuracy. For the thermal camera footage, a simple threshold-based method is shown to give useful detection performance. A Kalman filter is used to perform tracking based on the detection result. In parallel with obtaining the object position in images, the LIDAR output is used to convert the UAV's position from pixel space to global space. 
To train and validate the pipeline, we design a series of experiments in a motion capture environment that provides highly accurate global position data for the UAV. Additional public databases are used to augment the training process. The performance of our computational pipeline is evaluated in terms of position estimation error and computing cost.
    
The remainder of this paper is organized as follows: Section II outlines a survey of related work on vision-based object detection and tracking and their applications in the UAV domain. Section III describes the structure of our computational pipeline. Section IV describes the design of the experiment and discusses the results. The concluding remarks and plans for future work are briefly discussed in Section V.


\begin{figure*}[htbp]
    \centering
    \includegraphics[width=0.8\linewidth]{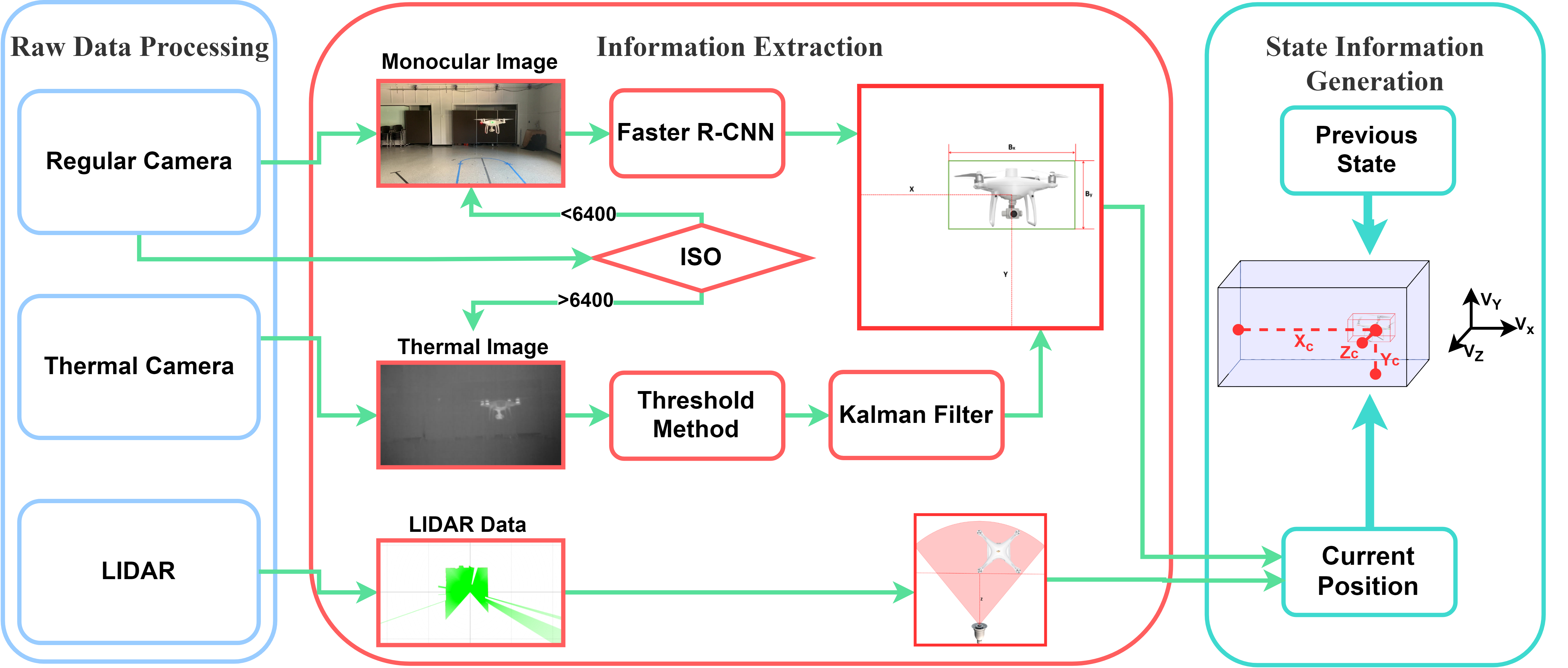}
    \caption{Computational Pipeline}
    \label{fig:pipeline}
\end{figure*}
\section{Literature Survey}
\subsection{Object Detection and Tracking}
Object detection is the primary step for object tracking. Every tracking method needs detection method to either provide an initial bounding box when the object first appears in its FoV (Field of View) or provide interest regions in every frame\cite{joshi2012survey}. Traditional object detection methods mainly rely on the manually extracted feature from an image in which the object is present. Recently deep learning-based methods have enabled considerable improvement in object detection tasks. State-of-art deep learning-based methods can be mainly classified into region proposal based methods and regression/classification based methods. Region proposal based models, as a two-stage process, implement a rough scan of the whole frame first and follow by focusing on the region of interest\cite{zhao2019object}.
On the other hand, the one-stage method, a regression-based approach directly classifies the object. Both of these methods have their pros and cons. Region proposal based methods represented by R-CNN can achieve higher accuracy on the tasks while being computationally expensive because of their two-stage process \cite{girshick2014rich}. While the one-stage methods are faster, their accuracy is compromised, especially in small object detection \cite{redmon2016you} \cite{liu2016ssd}.

Object tracking algorithms provide position information from each frame in the video, which can help locate the object of interest \cite{joshi2012survey}, or even predict its immediate future trajectory. 
There are two object tracking methods that are noteworthy: generative modeling methods and discriminative modeling methods \cite{zhao2019object}. Generative modeling methods use rich information from frames; therefore, it is easier to get an accurate result from complex scenes. 
Since it does not consider the background information, it is not immune to the interference from the background information. Discriminative methods take background information into account, so they are more robust when the object is out-of-view and/or occluded. However, discriminative modeling methods highly rely on the training dataset \cite{fiaz2018tracking} \cite{luo2014multiple}. Most of the traditional tracking methods belong to the generative modeling method such as Kalman filtering method \cite{kalman}, particle filtering method \cite{particle2003}, and mean-shift method \cite{vojir2014robust}. 
Nowadays, the most popular methods have two branches: correlation filtering methods (CSK \cite{henriques2012exploiting}, KCF/DCF \cite{henriques2014high}, and CN \cite{danelljan2014adaptive} etc.) and deep-learning-based methods \cite{bolme2010visual,henriques2014high,nam2015mdnet,VOT_TPAMI,bertinetto2016fully,held2016learning}. Though they are fast and robust, they still rely on the object detection methods to feed them the initial bounding box. As for deep-learning-based methods, their most significant issue is the computational cost, which is a barrier to online tracking. Nevertheless, they show excellent performance on the accuracy of tracking. Furthermore, with deep-learning-based methods, we can implement tracking-by-detection to simplify our pipeline.

\subsection{UAV Detection and Tracking}
Several algorithms have been developed for detecting and tracking UAVs using the monocular camera. In UAV detection stage, the motion-based models are widely used. They can be broadly categorized into background subtraction methods and optical flow methods \cite{li2016multi}. Background subtraction methods can extract abiding pixels between each pair of adjacent frames. Next, they use the current frame to subtract those background pixels for identifying the object UAV \cite{li2016multi,stevendraft,7991302}. These background subtraction methods have excellent performance when background is relatively stable. However, if the background is dynamic, or if the UAV hovers at the same position for a long time, these methods may not be able to accomplish the task. Optical flow methods capture the UAV via the local motion vector \cite{ye2018deep}. When the video flow gets blurred, the accuracy decreases dramatically. Besides, the appearance-based model also plays an important role in previous researches \cite{rozantsev2016detecting}. Threshold-based methods work well when input frames are simple, but they are incapable of handling frames with complex conditions \cite{lee2018detection}. In order to consider more appearance information, template matching methods are used to increase the accuracy of recognition \cite{wu2017vision}. Nonetheless, template matching heavily depends on the selection of a good template and lacks generalizability. Currently, deep-learning-based methods used in UAV detection domain are restricted in using Convolution Neural Network as the classifier. This approach does not fully take advantage of deep learning. In UAV tracking stage, most of the works mentioned above use the Kalman filter to track the object, which is fast and robust. However, some parameters, like motion model, need to be set manually.

In our computational pipeline, we use Faster R-CNN to process the monocular image \cite{ren2015faster}. This track-by-detection method consists of a single step rather than a sequential two-step process of detection and then tracking. As it only focuses on the current frame, the dynamicity of the background becomes irrelevant. For the thermal image, we implement threshold-based detection and Kalman filter tracking method since the thermal image is relatively simple.

\section{Method}
\subsection{The Computation Pipeline}
The computational pipeline refers to the flight state extraction process that takes in the sensor data and estimates the flight state of the flying object (UAV) in view. As shown in Figure \ref{fig:pipeline}, the computational pipeline comprises of three steps: raw data processing, object position extraction, and flight state estimation. 

First, the bounding box of the flying object is obtained from the monocular camera and thermal camera pipelines.  The LIDAR depth footage provides information regarding the spatial distance of the flying object. Considering the lighting condition, we read ISO (the sensitivity to the light) information from the monocular camera and set the threshold to 6400. Under good lighting conditions (ISO < 6400) the monocular camera pipeline is executed. Otherwise, if the ISO value is higher than 6400, the thermal camera pipeline is put into action.

Next, the x and y coordinates of the object are extracted from the bounding box. However, these x, y coordinates are in image coordinate system rather than global coordinate system. The planar LIDAR data is used to calibrate and enable the transformation of the image-based coordinates into the global coordinates. 

In the final step, the velocity of the object is estimated by interpolating the change of coordinates between frames in image frame.

\subsection{Monocular Image}

Monocular image processing is widely used in many domains. The advantages of the monocular image are high resolution, sufficient information, and the availability of a variety of processing methods. We used Faster R-CNN as our tracker to detect UAV in every frame. 

\subsubsection{Faster R-CNN}
Faster R-CNN combines feature extraction, region proposal, bounding box regression, and classification in the same network. It improves the computational speed and keeps the high accuracy of detection \cite{ren2015faster}.

Feature extraction network uses Inception network to extract features from the input frame \cite{szegedy2016rethinking}. The fully-connected-layers and softmax layer are removed from the Inception network. Therefore, the output of the last layer is a higher-level feature map. Next, region proposal network (RPN), a convolutional network, predicts the proposal target area and computes bounding box regression from the given feature map. Then, the RoI (Region of Interest) pooling layer calculates the proposal feature maps after accepting the feature map and the proposed target areas. Finally, classification layers classify the proposal feature maps and refine the bounding box predictions.

Faster R-CNN is a weighted-sum multi-objective optimization problem, and its multi-task loss is defined by Eq. \ref{eq:loss}:

\begin{equation}
    \begin{aligned}
        L\left(\left\{p_{i}\right\},\left\{t_{i}\right\}\right) & =\frac{1}{N_{cls}} \sum_{i} L_{cls}\left(p_{i}, p_{i}^{*}\right) \\& +\lambda \frac{1}{N_{reg}} \sum_{i} p_{i}^{*} L_{reg}\left(t_{i}, t_{i}^{*}\right)
    \end{aligned}
    \label{eq:loss}
\end{equation}
in which, $i$ is the index number of anchor in mini-batch; $p_i$ is the possibility of anchor predicted as a target; $p^*_i$ represents GT tags; $t$ is the predicted bounding box; $t^*$ is the coordinate vector of ground truth corresponding to positive anchor. Classification loss ($L_{cls}$) calculates the softmax loss of classifying anchors as positive if they are negative and vice-versa. Regression loss ($L_{reg}$) is the loss of the positive anchor (bounding box) regression \cite{ren2015faster}.

\subsubsection{Training}

The feature extraction network is pretrained by Microsoft COCO database \cite{lin2014microsoft}. Our overall training dataset is a combination of online UAV images, videos, and frames from our own collected videos. We collected 500 images from our own recorded videos of UAV and 500 additional images from online resources. This dataset is divided into two sets. 700 images of the dataset constitute the training set, and the remaining 300 constitute the testing set.
We manually labeled the UAV with the bounding box using LabelImg \cite{labelimg}.

We use 50000 training steps with a batch size of 10, and the initial learning rate is set to 0.0002 with decay value set to 0.95. 

\subsubsection{Thermal Image}


In low light conditions such as fog or night conditions or conditions where the background is less distinguishable compared to the object of interest (e.g., white UAV flying over snowy background), the monocular image-based object detection and tracking may not be effective. Irrespective of the lighting conditions thermal cameras can capture the thermal energy distribution of the scene and provide an alternative modality that can be used in poor light conditions. In such a scenario, our computational pipeline selects the thermal imaging system as our primary modality to track the UAV. The thermal image can reliably detect the UAV since the heat signature of motors and processor of the UAV are distinct than the background owing to their relatively higher temperature compared to background temperature. We use a simple threshold method to determine the initial bounding box of UAV in thermal images.  

We use the maximum correlation criterion based threshold method to find the segmentation threshold \cite{366472}. First, the distribution of the probability of each gray-level can be obtained from the input frame. Next, the maximum correlation between the object and background is treated as the criterion to choose the threshold. Finally, the frame is segmented into object and background. 

After getting the initial bounding box of the object, we apply an object tracking approach based on the Kalman filter. 


\subsection{LIDAR Data}
The observation value of the LIDAR sensor is the real distance between the sensor and the object. One big advantage of this modality is the minimal amount of computation required for distance computation. The distance computed by LIDAR data is used for reliable transformation of data from image coordinate to global coordinate system. Since the used LIDAR Sensor works in the 2D plane, the LIDAR pipeline is only executed when the UAV flies in the working plane of the LIDAR sensor.  

\begin{figure}
    \centering
    \includegraphics[width=0.9\linewidth]{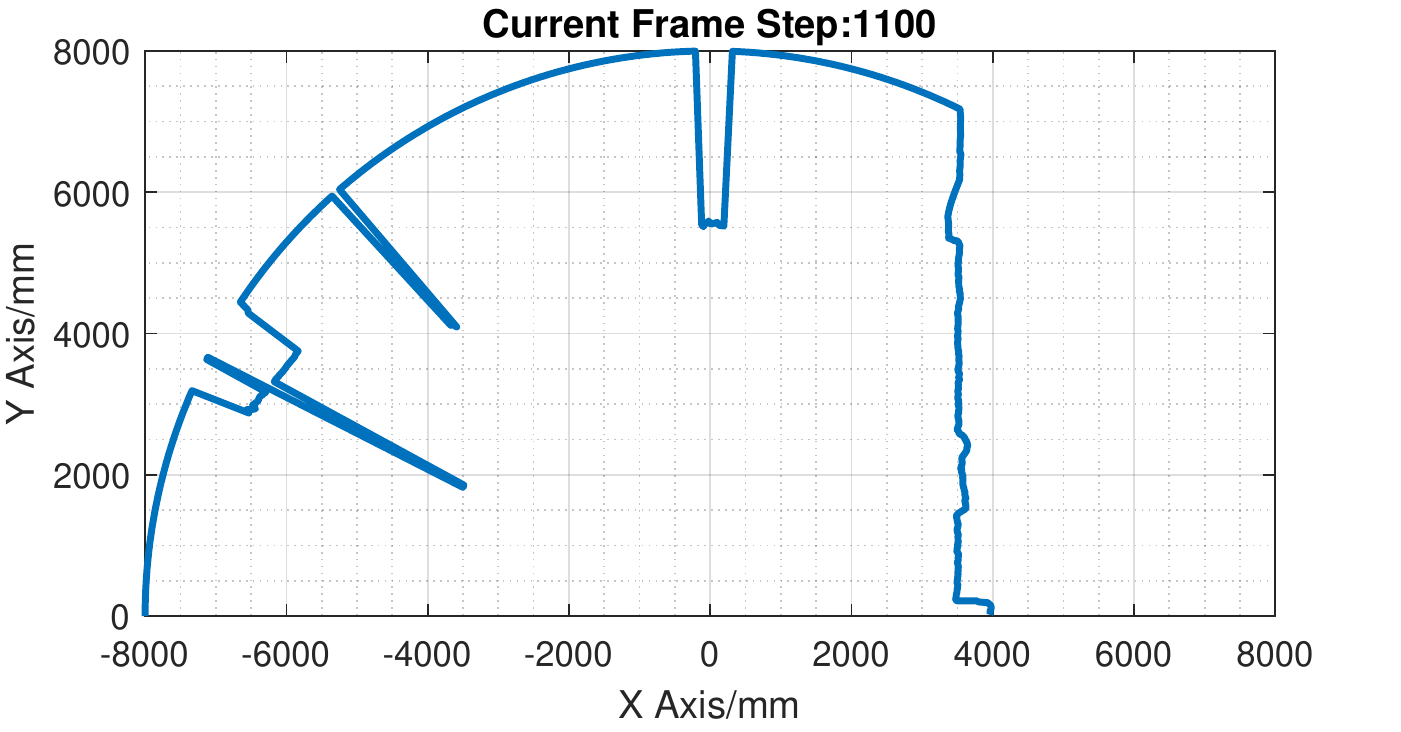}
    \caption{Example of LIDAR Map}
    \label{fig:lidarl}
\end{figure}

For our sensor, the detection range is 180 degree, and there are four laser detection points within each degree. After reading the distance, the possible objects can be found via the distance curve. In Figure \ref{fig:lidarl}, $X$ range between 3500 and 8000 is the wall of the room. Additional objects detected in the scene are filtered out. Besides, some noise in the distance curve is also observed sometimes. So when the detected object is smaller than 10 detection points, the objects in the scene are ignored and filtered out. We find the real distance $D$ and real length of object cutting area  $L$ by Equation \ref{eq:lidar}
    
\begin{equation}
    D = \frac{\sum_{i=1}^{N_p}d_i}{N_p} \qquad L = 2D\tan(\frac{N_p}{4})
    \label{eq:lidar}
\end{equation}
in which, $N_p$ is the number of active points, $d_i$ is the distance between each sensor detection point and the object.

\subsection{State Information Extraction}
From monocular camera and thermal camera pipeline, we obtain the bounding box of the UAV. We define the center of the frame as the origin of the image-based coordinate so that the central point of the UAV bounding box can be calculated. For reducing the computational complexity, we ignore the variation of the bounding box due to UAV's rotation (yaw, pitch, roll). Then we obtain the coordinate based on real measurement for monocular image and thermal image by Equation \ref{eq:coor}.

\begin{equation}
    C_R = C_P\frac{L}{l}
    \label{eq:coor}
\end{equation}
Here, $C_R$ is the image-based coordinate given by real measurement, $C_P$ is the image-based coordinate estimated from pixel measurement, $l$ is the length of the bounding box based on pixel measurement and $L$ represents the real length of the object derived from Equation \ref{eq:lidar}. After calibrating the measurement, the ratio between the current real distance $D_i$ and the calibrating distance $D_c$ is shown in Equation \ref{eq:distance},  

\begin{equation}
    \frac{D_c}{D_i} = \dfrac{\frac{F_c}{\alpha}}{\frac{F_i}{\alpha}}=\dfrac{\frac{f L_c}{l_c \alpha}}{\frac{f L_c}{l_i \alpha}} = \dfrac{l_i}{l_c}
    \label{eq:distance}
\end{equation}
in which, $F_i$ is current frame length as measured, $F_c$ is calibrated frame length, $f$ is the total pixel length of the frame, $l_i$ is current object length in pixel and $l_c$ is the calibrated object length in pixels. Using this expression, we can find the current distance $D_i$.

Finally, based on the real position information $p_t$, we can calculate the velocity of the UAV $v_t$ by Equation \ref{eq:v}
\begin{equation}
    v_t = \frac{p_t-p_{t-1}}{\Delta t}
    \label{eq:v}
\end{equation}

\section{Experiments}
\subsection{Hardware Setup and Data collection}
\begin{figure}[htbp]
\begin{center}
  \includegraphics[width=0.9\linewidth]{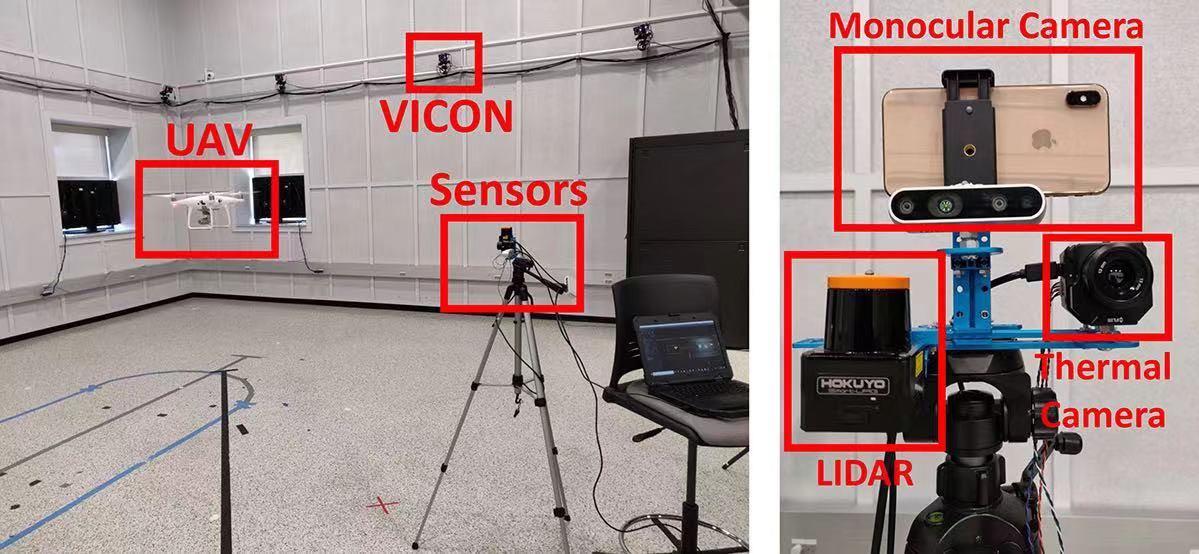}
  \caption{Hardware Setup. Left figure is our experiment environment in VICON system, right figure is the sensors we used in expeirment.}
  \label{fig:LTS}
\end{center}
\end{figure}

As shown in Figure \ref{fig:pipeline}, the integrated framework has three different modes of sensor input. The sensor information processing framework estimates the states of the object of interest from the raw data. Therefore a data pool is created in a controlled environment to build the computational pipeline. A simple design of the experiment is crafted where three different sensors are used to gather data of a flying quadcopter from a wide variety of height and distance w.r.t. the sensors. Sensors used in the experiment are shown in Figure \ref{fig:LTS} (right). The detailed description of the sensors is given in Table \ref{sensor_spec}. Serving as the object of interest, a DJI Phantom 4 Pro quadcopter is used in the VICON motion capture systems (as shown in Figure \ref{fig:LTS} left). All the sensors and the quadcopter are equipped with reflective markers to trace their exact location. The sensors have different output data rates (as shown in Table \ref{sensor_spec}); therefore, post-processing is used to filter and synchronize the data. Totally four sets of data from different sensors are collected, with the time span of the experiments varying from 5 minutes to 15 minutes. 

\begin{table}[htbp]
\begin{center}
\caption{Specification of Sensors}
\label{sensor_spec}
\begin{tabular}{lccc}
    \toprule
\textbf{Type of Sensor} & \textbf{Model} & \textbf{Specifications}\\
    \midrule
Monocular Camera & Iphone XS Max & 1920 x 1080 at 60 fps\\
    \midrule
 & & Detection Distance : 30m\\
 \multirow{2}{*}{LIDAR}& \multirow{2}{*}{Hokuyo ust-10lx} & Scan angle and speed : \\
 & & 180$^{\circ}$ at 40 Hz \\
 & & Accuracy: $\pm$ 40mm \\
    \midrule
Thermal Camera & FLIR Vue Pro 640 & 640 x 512 at 30 fps\\
    \midrule
 Exact loaction & \multirow{2}{*}{VICON System} & \multirow{2}{*}{3D coordinates at 100 Hz}\\
 (ground truth) & & \\

    \bottomrule
\end{tabular}
\end{center}
\end{table}

\subsection{Result of UAV Detection and Tracking}
\subsubsection{Monocular Results}
\begin{figure}[htbp]
\centering
    \subfigure[]{\includegraphics[width=0.235\textwidth]{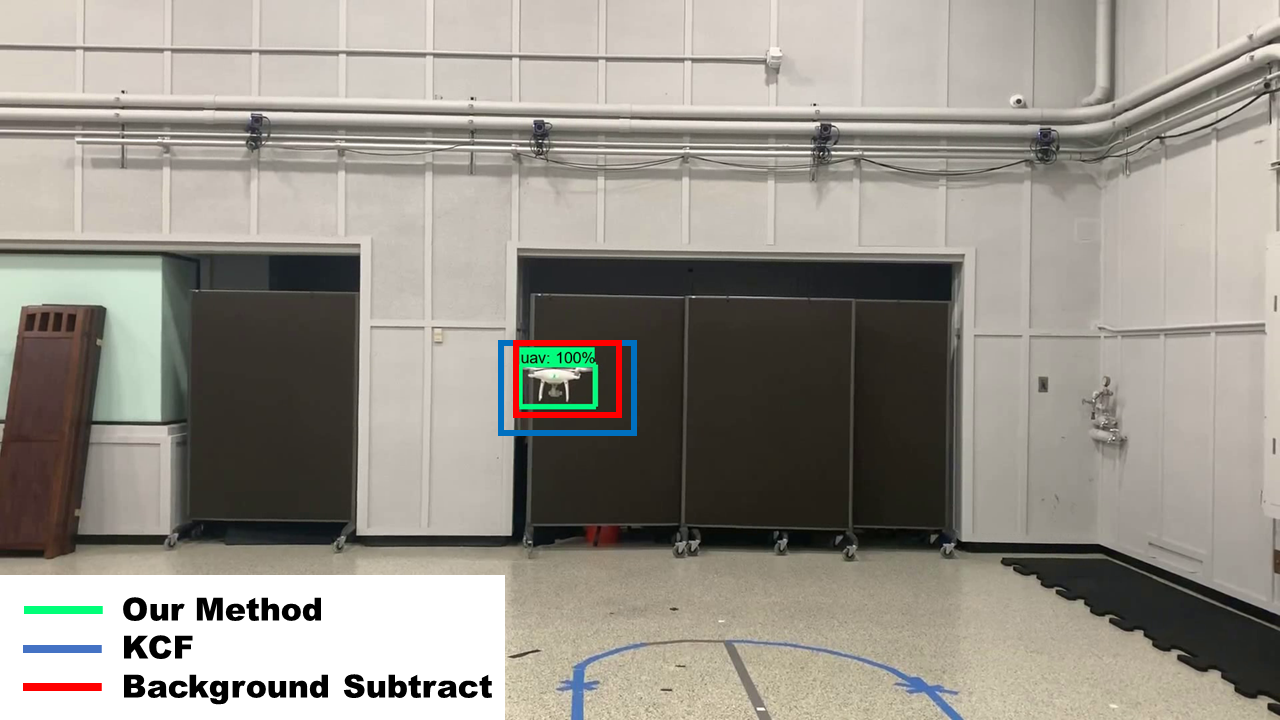}} 
    \subfigure[]{\includegraphics[width=0.235\textwidth]{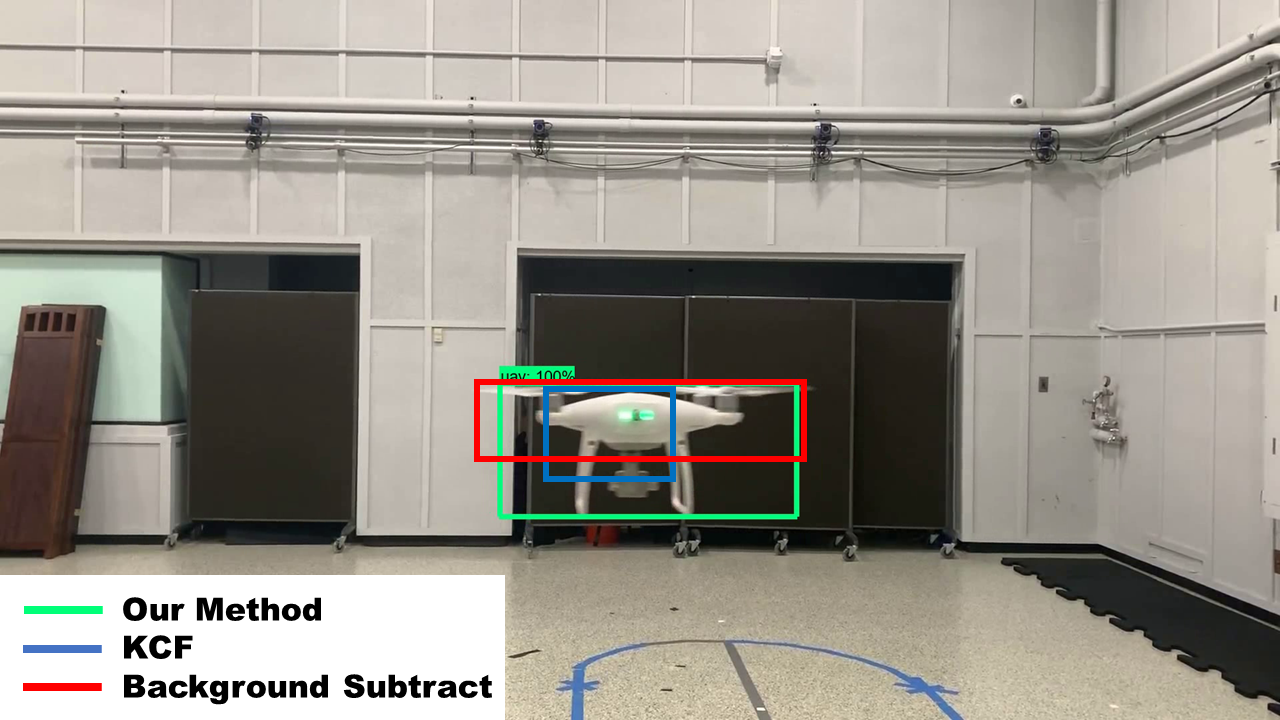}}
    \subfigure[]{\includegraphics[width=0.235\textwidth]{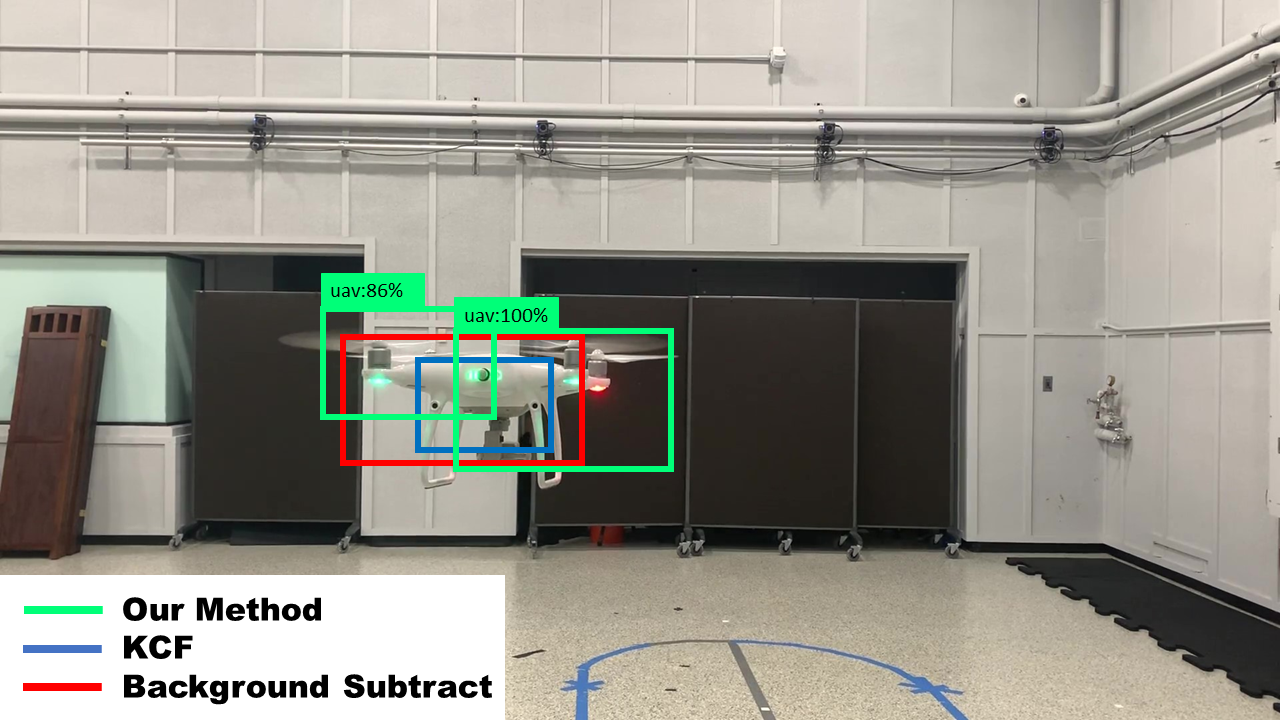}}
    \subfigure[]{\includegraphics[width=0.235\textwidth]{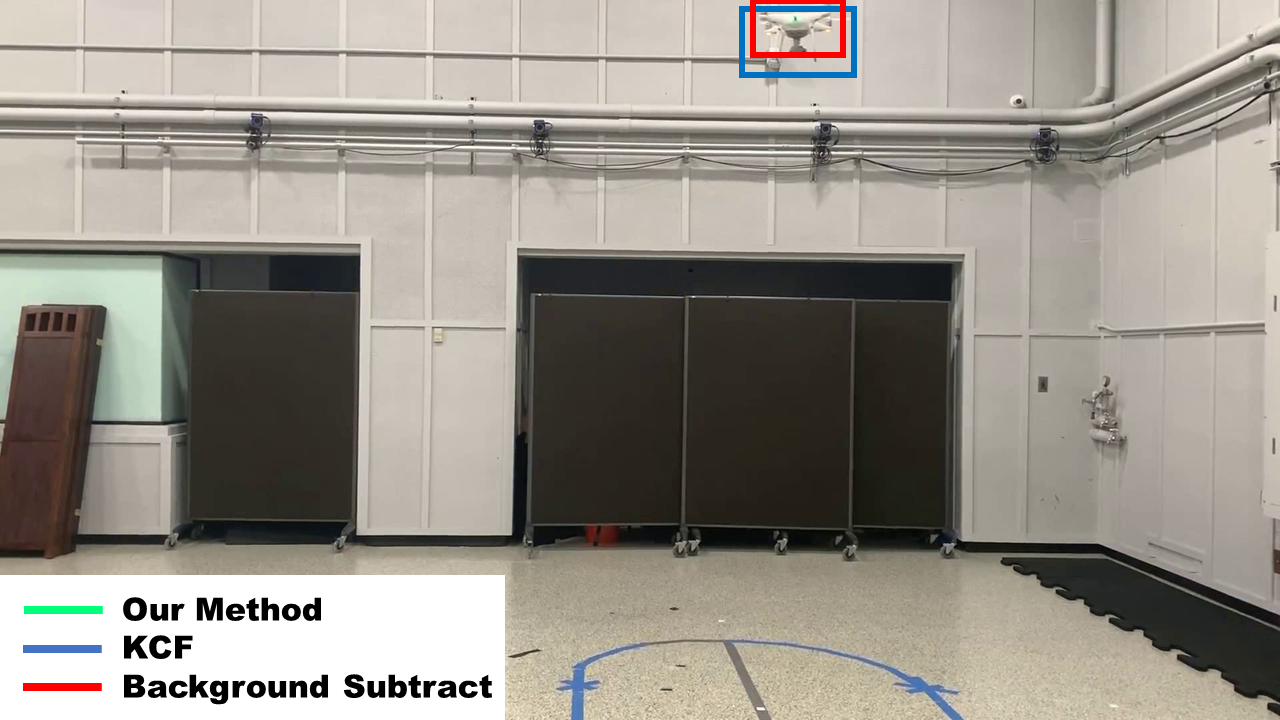}}
    \caption{Results Bounding Box Comparison For Monocular Image Between Different Methods}
\label{fig:mono_result}
\end{figure}

\begin{table}[]
\caption{Computational Resouce}
\label{tab:pc}
\begin{tabular}{ccccc}
\hline
 & \textbf{CPU} & \textbf{RAM} & \textbf{GPU} & \textbf{\begin{tabular}[c]{@{}c@{}}Operating \\ System\end{tabular}} \\ \hline
Training & \begin{tabular}[c]{@{}c@{}}Intel \\ Xeon 6130\end{tabular} & 128GB & \begin{tabular}[c]{@{}c@{}}NVIDIA \\ Tesla V100\end{tabular} & Centos 7.5.x \\ \hline
\multicolumn{1}{l}{Testing} & \begin{tabular}[c]{@{}c@{}}AMD \\ Ryzen 3700X\end{tabular} & 16GB & \begin{tabular}[c]{@{}c@{}}NVIDIA \\ RTX 2070 Super\end{tabular} & Windows 10 \\ \hline
\end{tabular}
\end{table}
Figure \ref{fig:mono_result} shows the comparison between our methods and popularly used methods: KCF tracking and background subtract method. In Figure \ref{fig:mono_result}(a)(b) we can observe that when UAV moves forward and backward, all three trackers can follow the object. However, the KCF and background subtraction methods incur a comparatively larger error in finding the bounding box than our method. These errors accumulate when we calculate the state information in the next step. In our experiment, KCF lost the object after 4 minutes, and KCF tracker can not follow the object when the object goes out of the frame and comes back again. Since this tracker is a short-term tracking method, it needs to be updated by the detection method after running for a specific time. In addition, when the UAV hovered for a while, the background subtraction method failed to find the object. We also observed two issues with our method. Since our method finds the bounding box that has a score over 80, when the object is too big in the frame, we may have multiple candidates bounding boxes with a score over 80$\%$. Secondly, when the UAV is at a higher altitude, the detection ability is reduced. This limitation results from the dataset, in which we did not have enough data that capture UAV footage from an angle looking upward. These two issues can be resolved with an additional increase in the size and variety of the dataset.

The computational resources used for training and testing are shown in Table \ref{tab:pc}. The total training time is 5 hours and 26 minutes. The results of image testing set are shown in Table \ref{tab:mono_result}. In the table, average accuracy is the top one candidate bounding box score on each frame of the testing data. The correct rate is defined as the total detected objects divided by the real number of objects in the dataset. It can be observed that we accomplish a desirable accuracy of more than 90\%.
\begin{table}[hpt]
\caption{Result of Faster R-CNN method and threshold-Kalman method on the testing dataset}
\begin{tabular}{cccc}
\hline
 & \textbf{Avg. Accuracy} & \textbf{Correct Rate} & \textbf{Avg. Speed} \\ \hline
Faster R-CNN & 99\% & 92\% & 17 fps \\ \hline
Thermal Threshold & \multirow{2}{*}{-} & \multirow{2}{*}{86\%} & \multirow{2}{*}{136 fps} \\ 
+Kalman &  &  &  \\ \hline
\end{tabular}
\label{tab:mono_result}
\end{table}
\subsubsection{Thermal Results}
\begin{figure}
\centering
    \subfigure[]{\includegraphics[width=0.115\textwidth]{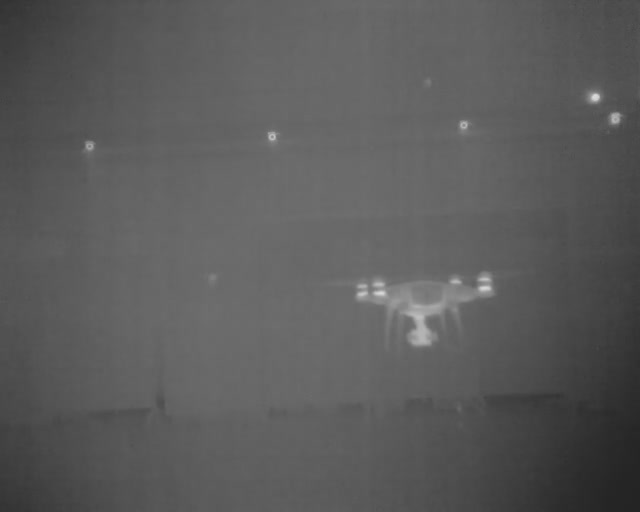}} 
    \subfigure[]{\includegraphics[width=0.115\textwidth]{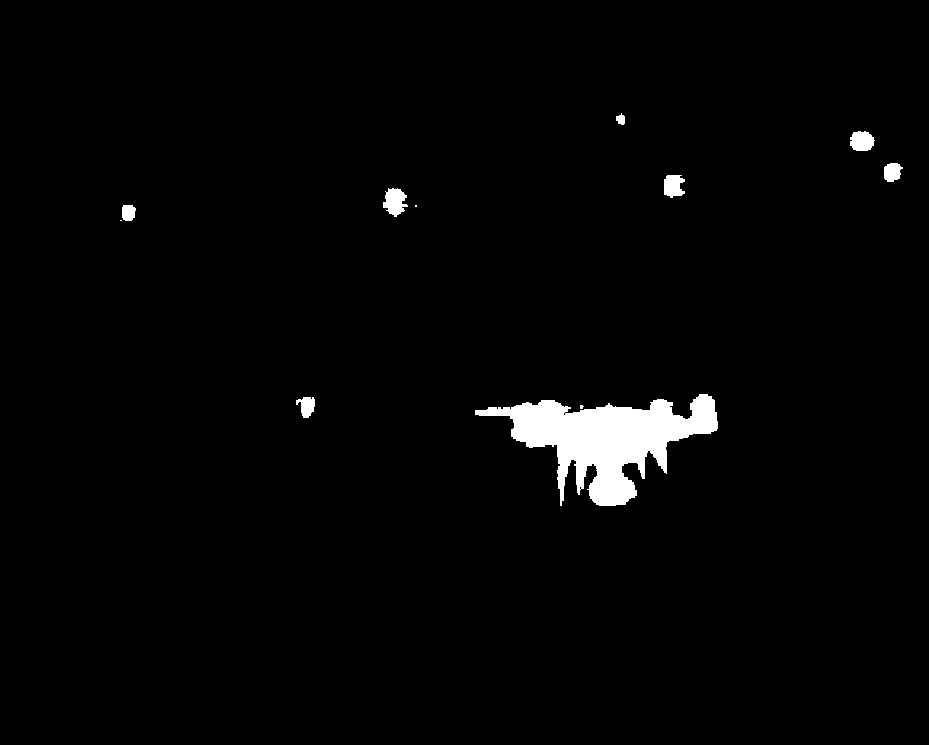}}
    \subfigure[]{\includegraphics[width=0.115\textwidth]{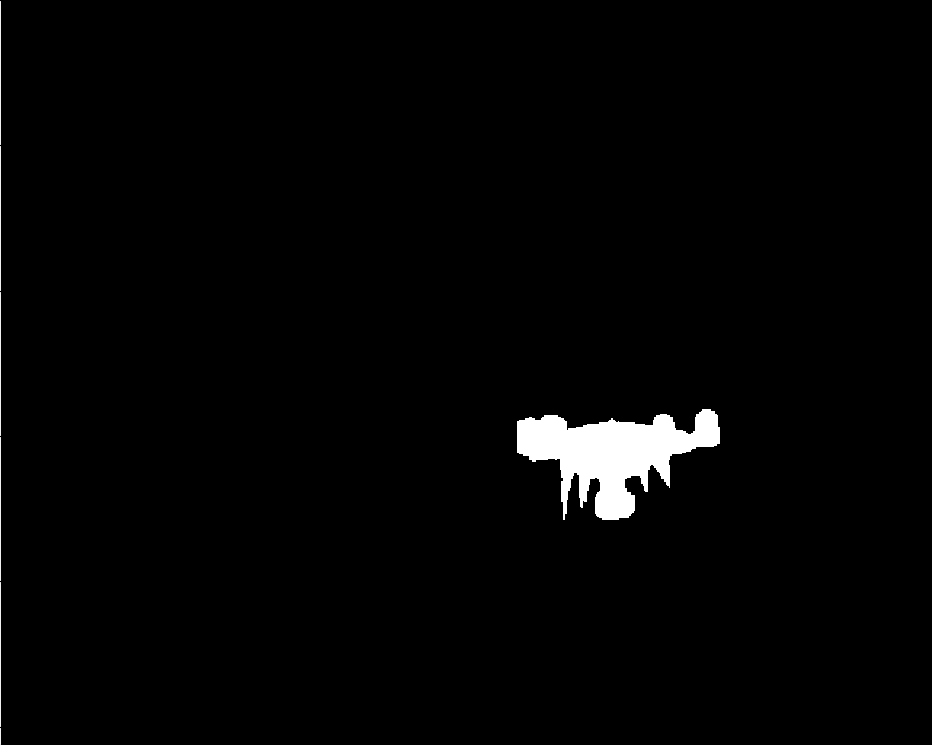}}
    \subfigure[]{\includegraphics[width=0.115\textwidth]{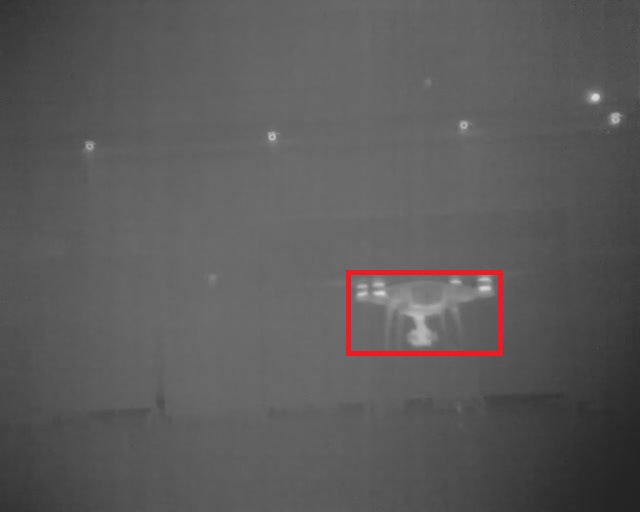}}
    \caption{(a) Original Thermal Image (b) Threshold Binary Image (c) Post-process Image (d) Tracking Result}
\label{fig:thermal_result}
\end{figure}

Figure \ref{fig:thermal_result} displays the intermediate steps of the thermal image tracking pipeline. The thermal image is a natural choice to get desirable tracking outcomes when thermal information of the environment is relatively simple. However, since our thermal camera changed to the ISO automatically, the threshold detection method failed in some cases. The results on testing data are shown in Table \ref{tab:mono_result} .

\begin{figure*}[t]
\centering
    \subfigure[]{\includegraphics[width=0.3\textwidth]{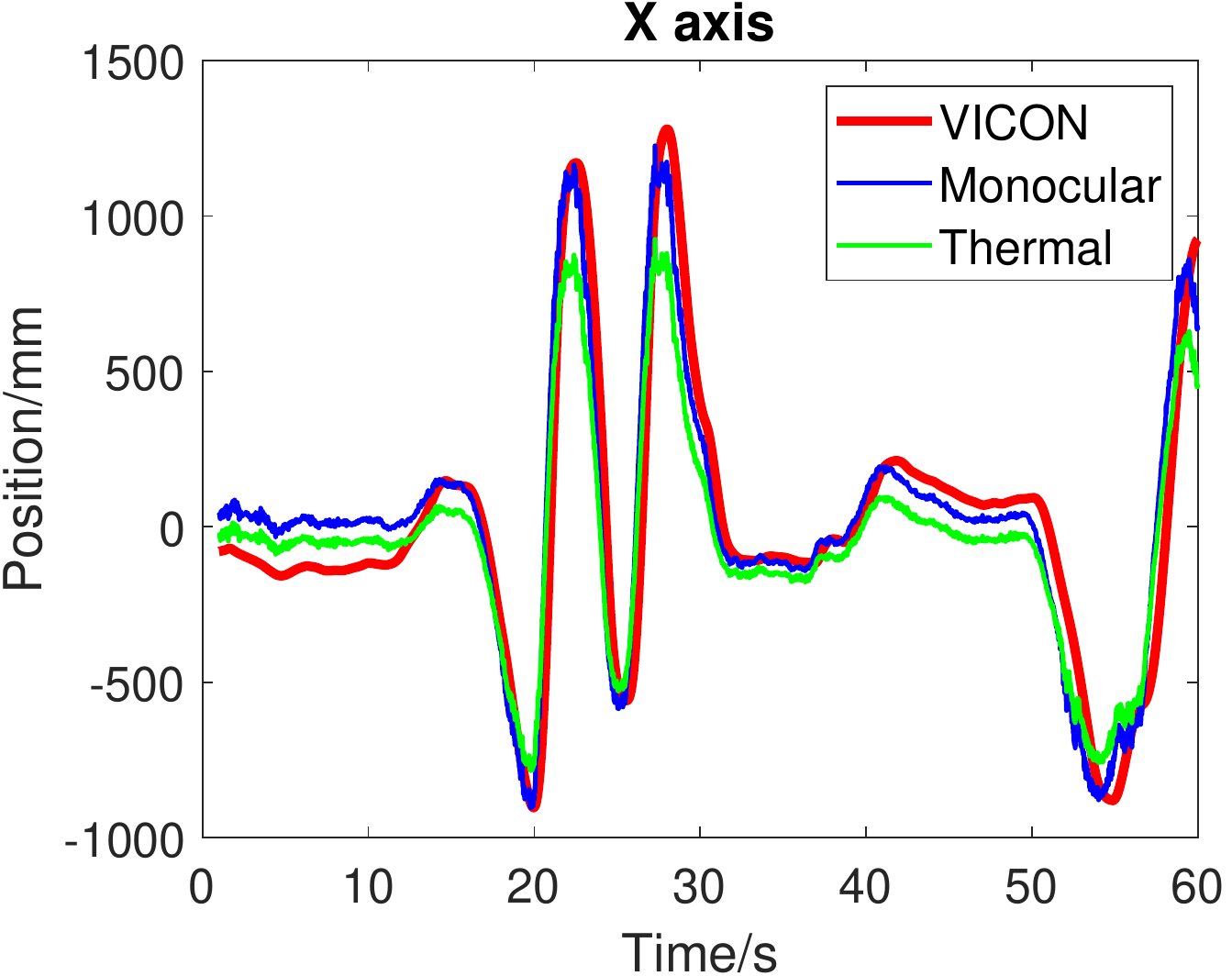}} 
    \subfigure[]{\includegraphics[width=0.3\textwidth]{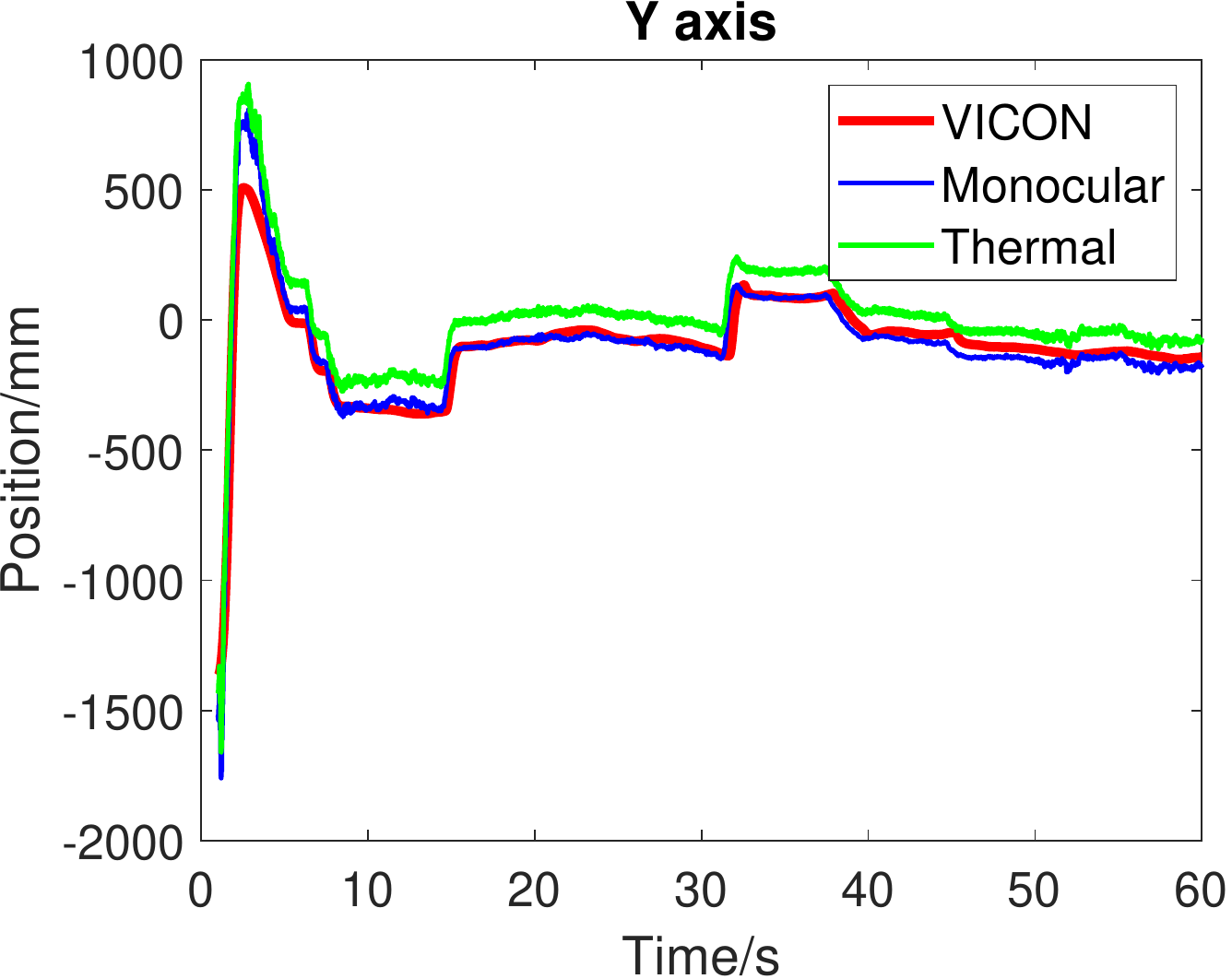}}
    \subfigure[]{\includegraphics[width=0.3\textwidth]{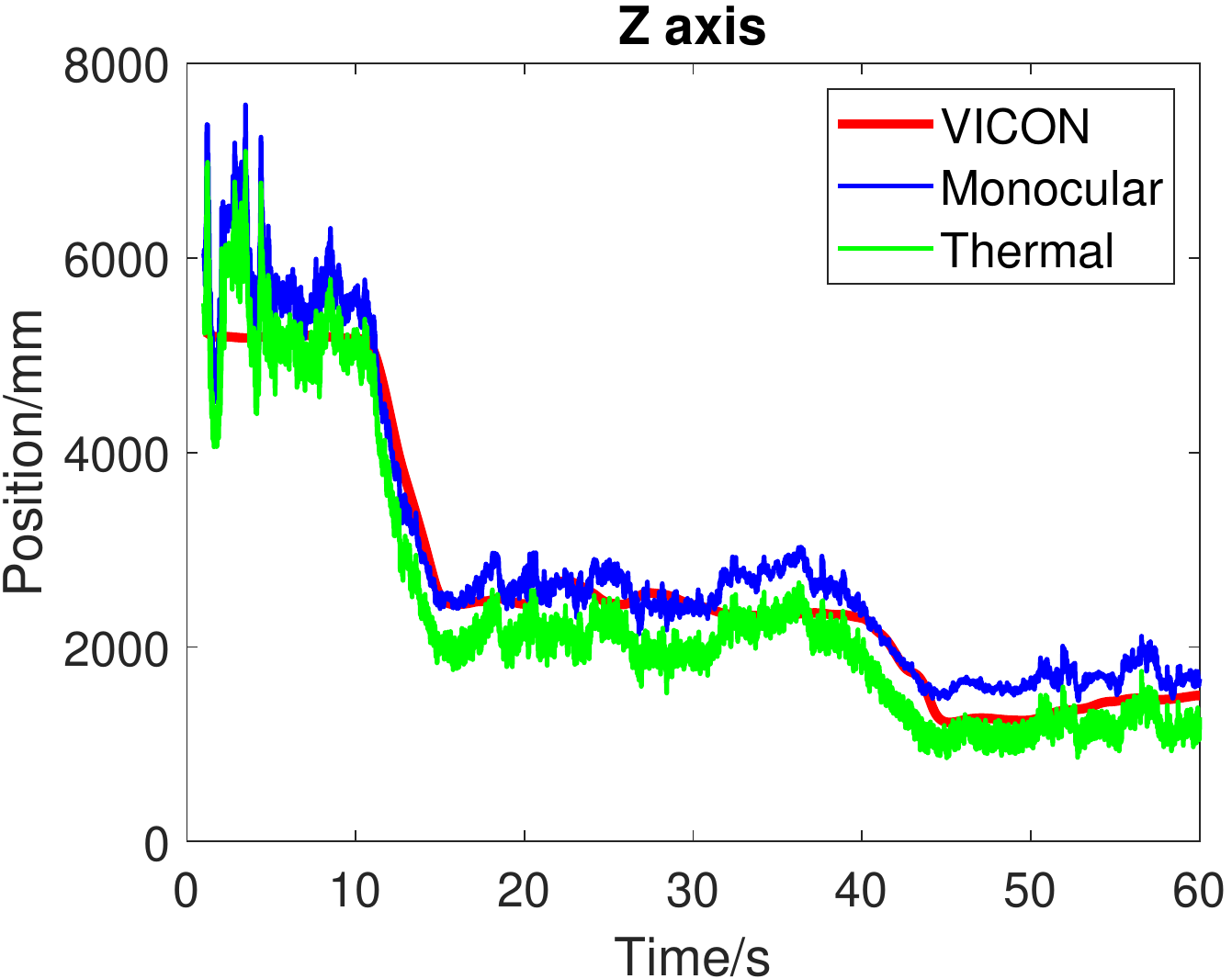}}
    \caption{Position Results on Testing Data for the Sample Experiment (a) Results in $x$ axis (b) Results in $y$ axis (c) Results in $z$ axis}
\label{fig:comparison}
\end{figure*}

\begin{figure}[]
\begin{center}
  \includegraphics[width=0.95\linewidth]{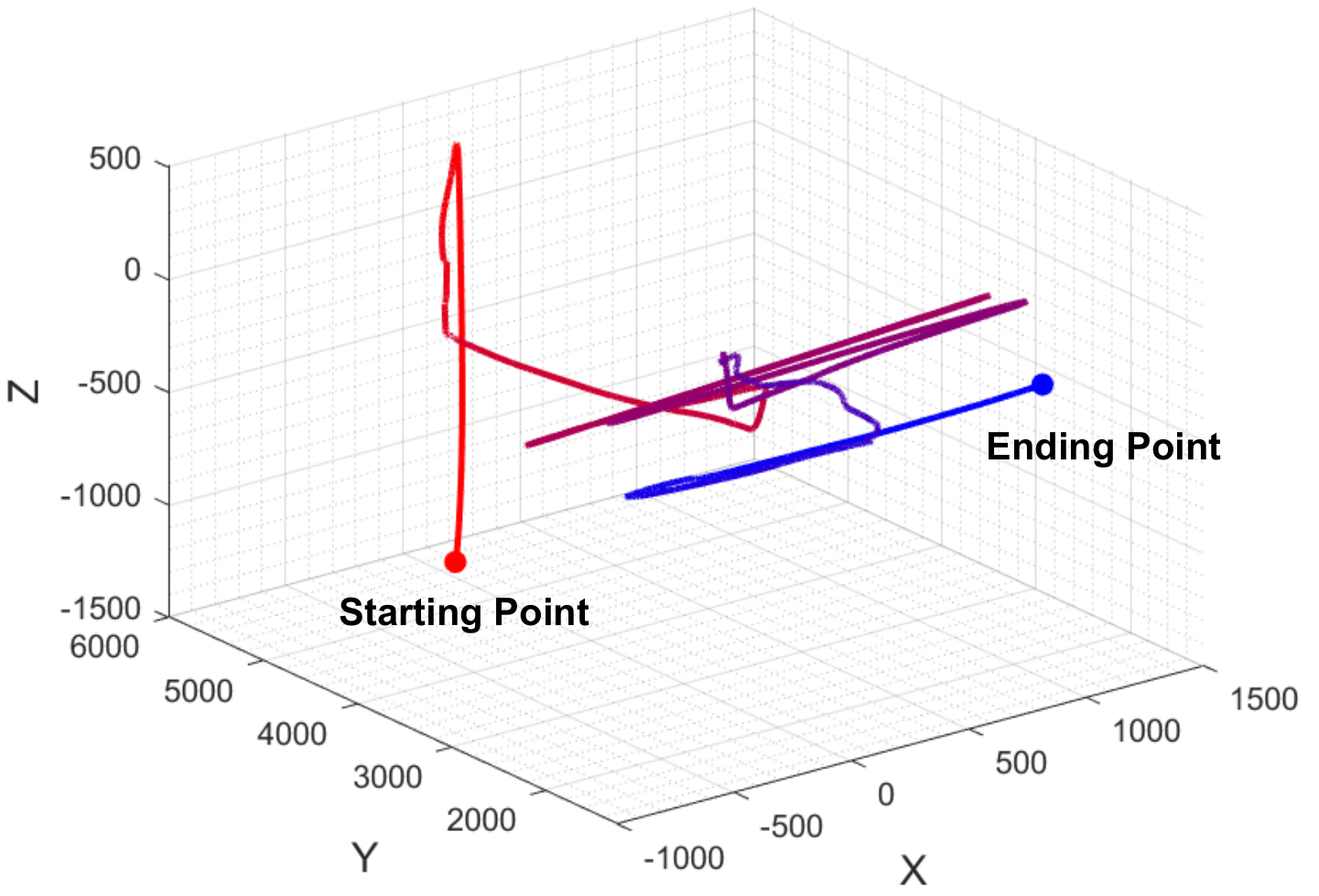}
  \caption{Ground truth trajectory of the UAV in the Sample Experiment}
  \label{fig:traj}
\end{center}
\end{figure}

\subsection{Quantitative Evaluation}
Here we use one set of videos as a sample experiment to demonstrate the performance of our detection/tracking/state extraction pipeline. All three sensor data and the VICON motion capture data were manually synchronized and cut for a length of 60s. Next, we run our pipeline to generate the image-based coordinate with real measurement. As shown in Figure \ref{fig:comparison}, the red line is the VICON data, the blue line is the result of monocular pipeline, and the green line is the result of thermal pipeline. VICON data is regarded as the ground truth, with the true trajectory for the sample experiment shown in Figure \ref{fig:traj}. As depicted in Figure \ref{fig:comparison}, our pipeline shows a promising performance on the experimental data. In the $Y$ axis, the result of thermal pipeline has an offset from the ground truth. Since, in the thermal image, the landing gear of UAV is relatively cool, and the motors are warmer, the center of captured bounding box is higher than that in the monocular pipeline. Lastly, in $Z$ axis, we observe notable noise in both monocular and thermal pipeline outcomes, w.r.t. the ground truth. This noise can be attributed to the error propagation caused by the pixel to distance conversion process (to be addressed in the future through multiple calibration runs).

Since different sensors have different sampling frequency, we applied linear interpolation to generate the continuous ground truth curve from VICON data. Next, we compared the positional error of the two pipelines (Table \ref{tab:position_result}). The percentage error in the table refers to the ratio of mean absolute error to the bounding box size of the UAV trajectory.

\begin{table}[]
\begin{center}

\caption{Position Result}
\label{tab:position_result}
\begin{tabular}{ccccccc}
\hline
 & \multicolumn{3}{c}{\textbf{Mean Absolute Error}} & \multicolumn{3}{c}{\textbf{Percentage Error}} \\
 & \textbf{X} & \textbf{Y} & \textbf{Z} & \textbf{X} & \textbf{Y} & \textbf{Z} \\ \hline
Monocular & 112.1 & 34.5 & 295.7 & 5.1\% & 1.8\% & 7.4\% \\ \hline
\multicolumn{1}{l}{Thermal} & \multicolumn{1}{l}{147.7} & \multicolumn{1}{l}{99.8} & \multicolumn{1}{l}{409.6} & \multicolumn{1}{l}{6.8\%} & \multicolumn{1}{l}{5.3\%} & \multicolumn{1}{l}{10.2\%} \\ \hline
\end{tabular}
\end{center}
\end{table}

\section{Concluding Remarks}
This paper develops and demonstrates a pipeline for integrating different sensors to improve the performance of tracking and localizing a flying UAV. The experimental results show that the error in the estimated position is within 10\%, with the tendency of motion well captured and consistent with the ground truth data. It is evident from our experimental results that the multi-sensor computation pipeline is capable of accurately and efficiently extracting the localization information about an approaching UAV. Compared to some state-of-the-art UAV tracking methods, our monocular footage process in the pipeline gives out more accurate bounding box. Moreover, utilizing a deep learning based approach entails doing away with the need to create a motion model of the incoming UAV and manual extraction of features. The use of the thermal camera grants the pipeline the ability to work in low light conditions. The final output of the pipeline is global coordinate positions that can be directly used for collision avoidance applications. 

Limited by the spatial extent of the indoor motion capture environment and the current data collection setup, the experiment dataset was restricted to a small range of motion and maneuvers. Extensive indoor and outdoor field experiments are planned in the future to extend this data set and test the online tracking performance of the pipeline in scenarios closer to real-life applications. Additionally, we will work on simplifying the Faster R-CNN network to speed up the regular footage processing and introduce a learning-based method to process the thermal footage with greater robustness.

\section*{Acknowledgement}
\noindent Support from the DARPA Award HR00111890037 from Physics of AI (PAI) program is gratefully acknowledged. 

\bibliographystyle{IEEEtran}
\bibliography{reference}

\end{document}